\documentclass[runningheads]{llncs}
\usepackage{graphicx}

\usepackage{tikz}
\usepackage{comment}
\usepackage{amsmath,amssymb}
\usepackage{color}

\usepackage{booktabs}
\usepackage{bbding}
\usepackage{xspace}
\usepackage{interval}
\usepackage{bm}
\usepackage{enumitem}
\usepackage{mathtools}
\usepackage{footmisc}
\usepackage{tabularx,booktabs}
\usepackage{multirow}

\usepackage{siunitx}

\usepackage{caption} 
\usepackage[accsupp]{axessibility}

\newcommand{\system}{STTS\xspace}
\newcommand*{\eg}{\emph{e.g.}\@\xspace}
\newcommand*{\etc}{\emph{etc}\@\xspace}
\newcommand*{\ie}{\emph{i.e.}\@\xspace}

\makeatletter
\newcommand\figcaption{\def\@captype{figure}\caption}
\newcommand\tabcaption{\def\@captype{table}\caption}
\makeatother

\makeatletter
\def\thanks#1{\protected@xdef\@thanks{\@thanks
        \protect\footnotetext{#1}}}
\makeatother

\definecolor{citecolor}{RGB}{0, 113, 188}
\newcommand{\Drop}[1]{\textcolor{blue}{\footnotesize{$\downarrow$#1}}}

\newcommand{\light}[1]{\textcolor{gray}{#1}}
\usepackage{hyperref}
\hypersetup{pagebackref=true,breaklinks=true,letterpaper=true,colorlinks,bookmarks=false, citecolor=citecolor}

\usepackage{orcidlink}

\begin{document}
\pagestyle{headings}
\mainmatter
\def\ECCVSubNumber{3837}  
\newcommand\blfootnote[1]{%
  \begingroup
  \renewcommand\thefootnote{}\footnote{#1}%
  \addtocounter{footnote}{-1}%
  \endgroup
}

\title{Efficient Video Transformers with Spatial-Temporal Token Selection}

\author{Junke Wang$^{1,2*}$\orcidlink{0000-0001-8129-7333}, Xitong Yang$^{3*}$ \orcidlink{0000-0003-4372-241X}, Hengduo Li$^{3}$ \orcidlink{0000-0001-5314-6853}, Li Liu$^{4}$, \\ Zuxuan Wu$^{1,2\dagger}$ \orcidlink{0000-0002-8689-5807}, Yu-Gang Jiang$^{1,2}$ \orcidlink{0000-0002-1907-8567}
}

\institute{$^{1}$Shanghai Key Lab of Intell. Info. Processing, School of CS, Fudan University \\
$^{2}$Shanghai Collaborative Innovation Center on Intelligent Visual Computing
$^{3}$University of Maryland, $^{4}$BirenTech Research\\
}

\titlerunning{Efficient Video Transformers with Spatial-Temporal Token Selection} 
\authorrunning{J. Wang et. al} 

\maketitle

\begin{abstract}
\blfootnote{$^*$Equal contributions. $^{\dagger}$Corresponding author.}
Video transformers have achieved impressive results on major video recognition benchmarks, which however suffer from high computational cost. In this paper, we present \system, a token selection framework that dynamically selects a few informative tokens in both temporal and spatial dimensions conditioned on input video samples. Specifically, we formulate token selection as a ranking problem, which estimates the importance of each token through a lightweight scorer network and only those with top scores will be used for downstream evaluation. In the temporal dimension, we keep the frames that are most relevant to the action categories, while in the spatial dimension, we identify the most discriminative region in feature maps without affecting the spatial context used in a hierarchical way in most video transformers. Since the decision of token selection is non-differentiable, we employ a perturbed-maximum based differentiable Top-K operator for end-to-end training. We mainly conduct extensive experiments on Kinetics-400 with a recently introduced video transformer backbone, MViT. Our framework achieves similar results while requiring 20\% less computation. We also demonstrate our approach is generic for different transformer architectures and video datasets. Code is available at \href{https://github.com/wangjk666/STTS}{https://github.com/wangjk666/STTS}.

\end{abstract}

\section{Introduction}
\label{sec:intro}

\begin{figure}[t]
\centering
\includegraphics[width=0.95\linewidth]{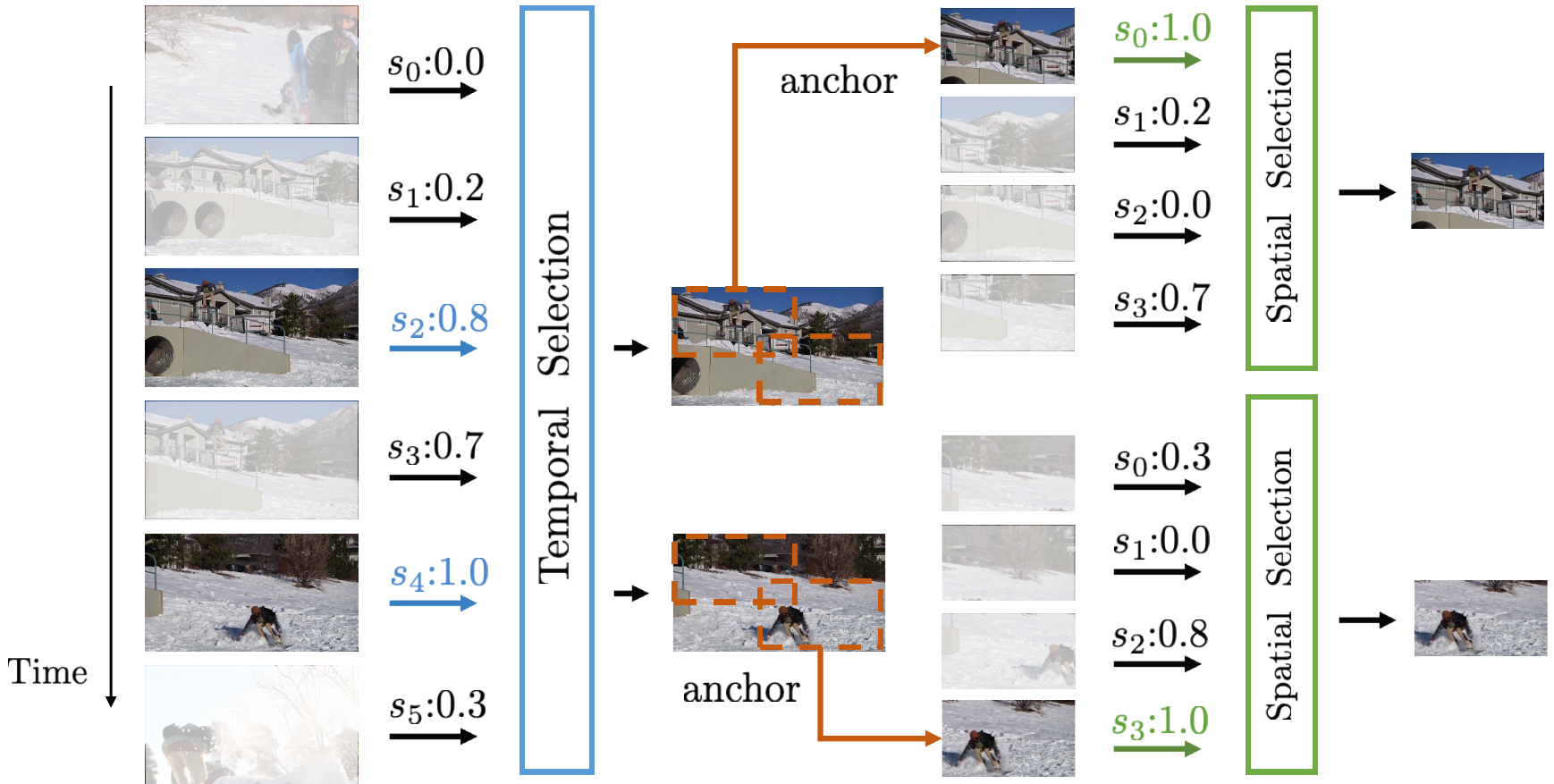}
\caption{A conceptual overview of our approach. We introduce a token selection method to improve the inference efficiency of video transformers by formulating token selection as a ranking problem. We use a lightweight scorer network to assign an importance score for each token, and only those with Top-K scores will be kept for computation. In the temporal dimension, we select the most informative frames, while in the spatial dimension, we preserve the most discriminative region with pre-defined anchors.}
\label{fig:teaser}
\end{figure}

The exponential growth of online videos has stimulated the research on video recognition, which classifies video content into actions and events for applications like content-based retrieval~\cite{dong2019dual,yuan2020central} and recommendation~\cite{davidson2010youtube,mei2011contextual,huang2016real,lee2017large}. At the core of video recognition is spatial-temporal modeling, which aims to learn how humans and objects move and interact with one another over time. 
Recently, vision transformers~\cite{dosovitskiy2021an,carion2020end} have attracted increasing attention due to their strong track record for capturing long-range dependencies in natural language processing (NLP) tasks~\cite{vaswani2017attention,devlin2018bert}. While achieving superior performance on major benchmarks, video transformers~\cite{gberta_2021_ICML,fan2021multiscale,liu2021video,wang2022bevt} are however computationally expensive. The problem stems from the fact that the number of input tokens grows linearly with respect to the number of frames in a clip, which further incurs a quadratic cost for computing self-attention. As a result, video transformers are oftentimes too compute-intensive to be deployed in resource-constrained scenarios.

While there are extensive studies on efficient video recognition for CNNs~\cite{zolfaghari2018eco,feichtenhofer2019slowfast,feichtenhofer2020x3d,tran2018closer,wu2019adaframe,korbar2019scsampler,wang2021adaptive,wang2021adafocus,li2021uav,sun2022human}, limited effort has been made for transformer-based video architectures. Unlike standard CNNs, transformers operate on image patches, which are then tokenized to a sequence of embeddings. The relationships among patches are modeled with stacked self-attention layers. In the image domain, a few very recent approaches have attempted to reduce the computational cost of transformers by learning to drop redundant tokens~\cite{rao2021dynamicvit,pan2021iared}, as transformers are shown to be resilient to patch drop behaviors~\cite{naseer2021intriguing}. 

Directly generalizing such an idea from image transformers to video transformers, while appealing, is non-trivial. Tokens in videos usually take the form of 3D cubes, and the tokenization layer in video transformers typically flattens all cubes into a sequence of 3D vectors. As a result, simply learning what to keep in the sequence with sampling based approaches~\cite{rao2021dynamicvit,pan2021iared} inevitably produces a set of \textit{discontinuous} tokens in space and time, thus destroying the structural information in videos. This also conflicts with the recent design of video transformers, which processes 3D tokens in a hierarchical manner preserving both spatial and temporal context~\cite{fan2021multiscale,liu2021video}.
Instead, we argue that the selection of spatial-temporal tokens in video transformers should be processed in a \textit{sequential} manner---attending to salient frames over the entire time horizon first, and then diving into those frames to look for the most important spatial region\footnote{Here the notion of ``frame'' can be either a single frame or multiple frames within a clip in the original video, depending on whether clip-based video models with 3D convolutions are used.}. 

In this paper, we introduce Spatial-Temporal Token Selection (\system), a lightweight and plug-and-play module that learns to allocate computational resources spatially and temporally in video transformers. In particular, \system consists of a temporal token selection network and a spatial token selection network, collaborating with each other to use as few tokens as possible. Each selection network is a multi-layer perceptron (MLP) that predicts the importance score of each token and can be attached to any location of a transformer model.  Conditioned on these scores, we choose a few tokens with higher scores for downstream processing. More specifically, given a sequence of input tokens, \system first selects a few important frames over the entire time horizon. Then for each frame, we split the token sequences into anchors with regular shapes, and select only one anchor which contributes most to video recognition. However, it is worth pointing out that selecting tokens with top scores is not differentiable and thus poses challenges for training. To mitigate this issue, we resort to a recently proposed differentiable Top-K selection algorithm~\cite{berthet2020learning} to make selection end-to-end trainable using the perturbed maximum method. This also allows us to explicitly control how many tokens are used. 

We conduct extensive experiments on two large-scale video datasets Kinetics-400~\cite{kay2017kinetics} and Something-Something-v2~\cite{goyal2017something} using MViT~\cite{fan2021multiscale} and VideoSwin~\cite{liu2021video} as backbones, to illustrate the effectiveness of our method. The experimental results demonstrate that \system can effectively improve the efficiency at the cost of only a slight loss of accuracy. In particular, by keeping only $50\%$ of the input tokens, \system reduces the computational cost measured in giga floating-point operations (GFLOPs) by more than $33\%$ with a drop of accuracy within $0.7\%$ on Kinetics-400 by using MViT~\cite{fan2021multiscale} as backbone. When more computational resources are allocated, \system achieves a similar performance as the original model but saves more than 20\% of the computation.

\section{Related Work}
\label{sec:related}
\noindent\textbf{Vision Transformers}
The great success of Transformer models~\cite{vaswani2017attention} in NLP has inspired the shift of backbone architectures from CNNs to transformers in the computer vision community~\cite{dosovitskiy2021an,zhang2020feature,touvron2021training,heo2021rethinking}. Vision transformers have demonstrated the capability to achieve state-of-the-art results in image domain, spanning a wide range of tasks like image classification~\cite{zhang2020feature}, object detection~\cite{zhu2021deformable}, semantic segmentation~\cite{zheng2021rethinking}, \etc,  with large-scale pre-training data.

Recently, a series of approaches~\cite{gberta_2021_ICML,fan2021multiscale,liu2021video,arnab2021vivit,gabeur2020multi,wang2021end,ryoo2021tokenlearner} explored the use of vision transformers in the video domain. TimeSformer~\cite{gberta_2021_ICML} adapts the standard transformer architecture to videos by concatenating the patches from different frames in the time dimension. MViT~\cite{fan2021multiscale} uses a hierarchical transformer architecture, which progressively expands the channel dimension and reduce the spatial resolution to extract the multiscale visual features for video recognition.  VideoSwin~\cite{liu2021video} extends the window-based local self-attention~\cite{liu2021swin} to video modeling by incorporating the inductive bias of locality. These approaches typically split the input video into spatial-temporal tokens to learn temporal relationships. While offering decent results, video transformers are computationally expensive. This motivates us to explore spatial and temporal redundancies in videos for efficient video recognition.
\newline \\
\textbf{Efficient Video Recognition}
Over the past few years, there are extensive studies on video recognition investigating the use of convolutional neural networks~\cite{feichtenhofer2016convolutional,hara2018can,feichtenhofer2019slowfast,xu2021sutd} and transformer models~\cite{gberta_2021_ICML,fan2021multiscale,liu2021video}. However, these models are usually computationally expensive, which spurs the development of efficient video recognition methods~\cite{yeung2016end,zolfaghari2018eco,wu2019adaframe,bhardwaj2019efficient,liu2020teinet,zheng2020dynamic,wang2021tdn,DBLP:journals/ijcv/WuLZXJD21} to speed up the inference time of video recognition. Several studies~\cite{zolfaghari2018eco,wu2018compressed,feichtenhofer2019slowfast,feichtenhofer2020x3d,kondratyuk2021movinets} explore designing lightweight models for video recognition by compressing 3D CNNs. Despite the significant memory footprint savings, they still need to attend to every temporal clip of the input video, thus bringing no gains in the computational complexity reduction. A few recent approaches~\cite{korbar2019scsampler,bhardwaj2019efficient,wu2019adaframe,wu2020dynamic}, on the other hand, propose to select the most salient temporal clips to input to backbone models for resource-efficient video recognition. Unlike these approaches that focus on the acceleration of CNN-based video recognition models, to the best of our knowledge, we are the first to explore efficient recognition for video transformers. It is also worth pointing out that \system is orthogonal and complementary to recent approaches on designing efficient vision transformers~\cite{wang2020linformer,kitaev2020reformer}.
\newline \\
\textbf{Differentiable Token Selection}
The decision of token selection is discrete, making it unsuitable for end-to-end training. To overcome this limitation, one solution is to resort Gumbel-Softmax trick~\cite{jang2016categorical} to decide whether each token should be pruned~\cite{pan2021iared,rao2021dynamicvit}. This however cannot explicitly control the number of preserved tokens during training, thus conflicting with the common hierarchical design of video transformers~\cite{fan2021multiscale,liu2021video,li2022uniformer}. Another way is to formulate the token selection as a ranking problem, in which the optimal-transport based methods~\cite{xie2020differentiable,cuturi2019differentiable} can be employed to match an auxiliary probability measure supported on increasing values. In this paper, we follow~\cite{cordonnier2021differentiable} to apply a perturbed maximum method, which is verified by~\cite{cordonnier2021differentiable} to outperform Sinkhorn operator~\cite{xie2020differentiable}.

\begin{figure*}[t]
  \centering
  \includegraphics[width=\linewidth]{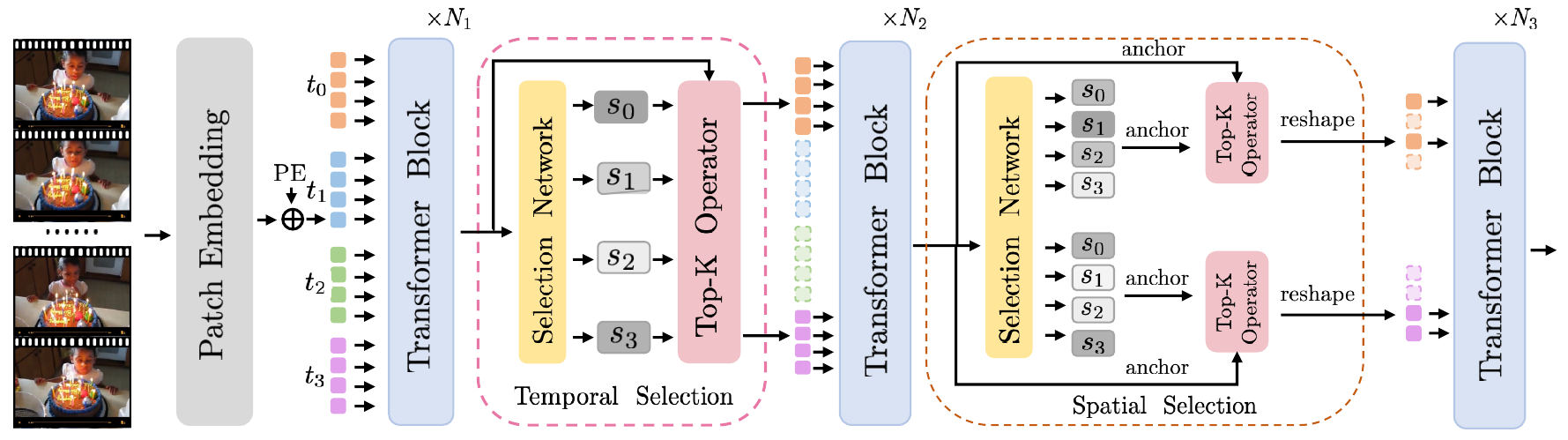}
  \caption{Overview of the proposed \system. The temporal token selection networks and spatial token selection networks can be inserted between the transformer blocks to perform token selection in the temporal and spatial dimension, respectively. $N_{1}$, $N_{2}$, $N_{3}$ could be 0. }
  \label{fig:network}
\end{figure*}

\section{Spatial-Temporal Token Selection}
\label{sec:method}

In this section, we begin with a brief review of video transformers (Sec.~\ref{sec:review}). Then we describe our token selection module that adaptively chooses a few important tokens, and introduce an important technique, the perturbed maximum method, for end-to-end optimization of the model (Sec.~\ref{sec:dtsn}). Finally, we elaborate how to instantiate the token selection modules in both temporal and spatial dimensions (Sec.~\ref{sec:tsts}). The overall framework is illustrated in Figure~\ref{fig:network}. 

\subsection{Review of Video Transformers}
\label{sec:review}
Given a video $\mathcal{V} \in \mathbb{R}^{T \times H \times W \times 3}$ consisting of $T$ RGB frames with the size of $H \times W$, video transformers typically adopt one of the two possible methods to map the video frames to a sequence of patch embeddings. The first is to tokenize 2D patches within each frame independently with 2D convolutions and concatenate all the tokens together along the time dimension~\cite{gberta_2021_ICML}, while the other is to directly extract 3D tubes from the input videos and use 3D convolutions to linearly project them into 3D embeddings~\cite{fan2021multiscale,liu2021video}. In both cases, the number of tokens is proportional to the temporal length and spatial resolution of the input video. We denote the resulting spatial-temporal patch embeddings as  $\mathbf{x} \in \mathbb{R}^{M \times N \times C}$, where $\textit{M}$ and $\textit{N}$ denote the length of the token sequence in time and space dimensions respectively, and $C$ is the embedding dimension. Positional encodings are added to $\mathbf{x}$ to inject location information~\cite{gberta_2021_ICML,fan2021multiscale,liu2021video}.

In order to model the appearance and motion cues in videos, the patch embeddings $\mathbf{x}$ are fed to a stack of transformer blocks which compute the spatial and temporal self-attention jointly~\cite{fan2021multiscale,liu2021video} or separately~\cite{arnab2021vivit,gberta_2021_ICML}. The self-attention is generally formulated as:
\begin{equation}
    Attention(\mathbf{Q}, \mathbf{K}, \mathbf{V}) = \texttt{softmax}(\frac{\mathbf{Q}\mathbf{K}^\mathrm{T}}{\sqrt{C}})\mathbf{V},
    \label{eq:self_att}
\end{equation}
where $\mathbf{Q}$, $\mathbf{K}$, $\mathbf{V}$ denote the query, key, and value embeddings based on $\mathbf{x}$, respectively, and $\texttt{softmax}$ denotes the normalization function.

\subsection{Dynamic Token Selection}
\label{sec:dtsn}
From Eqn.~\ref{eq:self_att}, we can see that the computational complexity of a video transformer grows quadratically with respect to the number of tokens used in the self-attention blocks. Considering the intrinsic spatial and temporal redundancies in videos, a nature way to reduce computation is to reduce the number of tokens. However, determining which tokens to be kept or discarded is non-trivial, which is closely related to both the input sample and the target task at hand. Inspired by the recent work on patch selection for high-resolution image recognition~\cite{cordonnier2021differentiable}, we formulate the token selection as a ranking problem---importance scores of input tokens are first estimated using a lightweight scorer network and the top $K$ scoring tokens are then selected for downstream processing. Below we introduce these two steps in details and present how to apply them to spatial and temporal token selection, respectively.

\paragraph{\bf Scorer network.} 
Given a sequence of input tokens $\mathbf{q} \in \mathbb{R}^{L \times C}$, the goal of the scorer network is to generate an input-conditioned importance score for each token. Here, $L$ denotes the flattened sequence length and $C$ is the embedding dimension.
We adopt a standard two-layer fully connected (FC) network to generate such scores.
In particular, the input tokens are first mapped to a \textit{local representation} $\mathbf{f^{l}}$ via a linear projection:
\begin{equation}
    \mathbf{f^{l}} = \texttt{FC} \left(\mathbf{q} ; \; {\bm {\theta}}_1 \right) \in \mathbb{R}^{L \times C^{\prime}}, 
\end{equation}
where $\bm{\theta}_1$ denotes the network weights and we use $C^{\prime}=C/2$ to save computation.
To leverage the contextual information of the whole sequence, we average $\mathbf{f^{l}}$ to get a \textit{global representation} $\mathbf{f^{g}}$ and concatenate it with each local representation along the channel dimension: $\mathbf{f_{i}} = \left[\mathbf{f^{l}_{i}}, \mathbf{f^{g}} \right] \in \mathbb{R}^{2C^{\prime}} (1 \leq i \leq L)$. The concatenated features are then fed to a second FC layer to generate the importance scores:
\begin{equation}
    \begin{split}
    \mathbf{s^{\prime}} &=  \texttt{FC} \left(
    \mathbf{f}; \; {\bm {\theta}}_2 \right)  \in \mathbb{R}^{L \times 1}, \\
    \mathbf{s} &= \frac{\mathbf{s^{\prime}} - \texttt{min}(\mathbf{s^{\prime}})}{\texttt{max}(\mathbf{s^{\prime}}) - \texttt{min}(\mathbf{s^{\prime}})},
    \end{split}
\end{equation}
where ${\bm {\theta}}_2$ is the network weights and $\mathbf{s}\in \mathbb{R}^{L}$ is the score vector of all tokens normalized with the min-max normalization~\cite{cordonnier2021differentiable}. 
It is worth mentioning that the additional computational overhead brought by the scorer network is negligible compared to the computation cost saved by pruning the uninformative tokens, as shown in Sec.~\ref{sec:results}.

\paragraph{\bf Differentiable Top-K Selection.}  Given the importance scores $\mathbf{s}$ generated from the scorer network, we select the $K$ highest scores and extract the corresponding tokens. We denote this process as a Top-K operator which returns the indices of the $K$ largest entries:
\begin{equation}
    \mathbf{y} = \texttt{Top-K}(\mathbf{s}) \in \mathbb{N}^{K}.
\end{equation}
We assume that the indices are sorted to preserve the order of the \textit{input sequence}. 
To implement token selection using matrix multiplication, we convert $\mathbf{y}$ into a stack of $K$ one-hot $L$-dimensional vectors $\mathbf{Y} = \left[ I_{y_{1}}, I_{y_{2}}, ..., I{y_{K}}\right] \in \left\{ 0,1\right\}^{L \times K}$. As a result, tokens with top $K$ scores can be extracted as $\mathbf{q^{\prime}} = \mathbf{Y}^\mathrm{T}\mathbf{q}$. 
Note that this operation is non-differentiable because both Top-K and one-hot operations are non-differentiable.

To learn the parameters of the scorer network using an end-to-end training without introducing any auxiliary losses, we resort to the perturbed maximum
method~\cite{berthet2020learning,cordonnier2021differentiable} to construct a differentiable Top-K operator. In particular, selecting Top-K tokens is equivalent to solving a linear program of the following form:
\begin{equation}
    \mathop{\arg \max}\limits_{\mathbf{Y} \in \mathcal{C}} \left \langle \mathbf{Y}, \mathbf{s1}^{T} \right \rangle,
\label{topk}
\end{equation}
where $\mathbf{s1}^{T}$ is the score vector $\mathbf{s}$ replicated $K$ times, and $\mathcal{C}$ is the convex polytope constrain set defined as:
\begin{equation}
\begin{split}
   \mathcal{C} = 
   \left\{ 
   \mathbf{Y} \in \mathbb{R}^{N \times K}: \mathbf{Y}_{n,k} \ge 0, \mathbf{1^{T}Y=1}, \mathbf{Y1 \le 1},   \right.\\
   \sum_{i \in \left[ N \right]} i \mathbf{Y}_{i,k}  < 
   \sum_{j \in \left[ N \right]} j \mathbf{Y}_{j,k^{\prime}}, \forall k < k^{\prime} &\left.
   \right\}.
\end{split}
\end{equation}
We follow~\cite{berthet2020learning} to perform forward and backward operations to solve Eqn.~\ref{topk}.

\textbf{-- Forward: } A smoothed version of the Top-K operation in Eqn.~\ref{topk} could be implemented by taking the expectation with respect to random perturbations:
\begin{equation}
    \mathbf{Y}_{\sigma} = \mathbb{E}_{Z} \left[ \mathop{\arg \max}\limits_{\mathbf{Y} \in \mathcal{C}} \left \langle \mathbf{Y}, \mathbf{s1^{T}} + \mathbf{\sigma Z} \right \rangle \right],
    \label{eq:sigma}
\end{equation}
where $\mathbf{Z}$ is a random noise vector sampled from the uniform Gaussian distribution and $\sigma$ is a hyper-parameter controlling the noise variance. In practice, we run the Top-K algorithm for \textit{n} (which is set to 500 in all our experiments) times, and compute the expectation of \textit{n} independent samples.  

\textbf{-- Backward: } Following~\cite{abernethy2016perturbation}, the Jacobian of the above forward pass can be calculated as:
\begin{equation}
    J_{\bf s}\mathbf{Y} = \mathbb{E}_{Z} \left[ \mathop{\arg \max}\limits_{\mathbf{Y} \in \mathcal{C}} \left \langle \mathbf{Y}, \mathbf{s1^{T}} + \mathbf{\sigma Z} \right \rangle \mathbf{Z^{T}} / \sigma \right].
\end{equation}
The equation above has been simplified in the special case where $\mathbf{Z}$ follows a normal distribution. With this, we can back-propagate through the Top-K operation. 

We train the backbone models together with the token selection networks using the cross-entropy loss in an end-to-end fashion. During inference, we leverage hard Top-K (implemented as torch.topk in Pytorch~\cite{paszke2019pytorch}) to further boost the efficiency. We follow~\cite{cordonnier2021differentiable}, where only a single Top-K operation will be performed (instead of n perturbed repetitions) and token selection is implemented with slicing the tensor. However, applying hard Top-K during inference results in a train-test gap. To bridge this, we linearly decay $\sigma$ to zero during training. Note that when $\sigma$ = 0, the differentiable Top-K operation is equivalent to hard Top-K and the gradients flowing into the scorer network vanish.

\subsection{Instantiations in Space and Time}
\label{sec:tsts}
Considering the distinct structure of appearance and motion cues in videos, we perform token selection \textit{separately} in space and time---attending to salient frames first and then diving into those frames to look for the most important spatial region. Similar idea has been explored in prior work for decoupling 3D convolutional kernels~\cite{tran2018closer}, spatio-temporal non-local blocks~\cite{he2020gta} and self-attention~\cite{gberta_2021_ICML}. 

\paragraph{\bf Temporal selection.} Given the input tokens $\mathbf{x}$ $\in$ $\mathbb{R}^{M \times N \times C}$, the goal of temporal selection is to select $K$ of the $M$ frames and discard the rest. We first average-pool $\mathbf{x}$ along the spatial dimension to get a sequence of frame-based tokens $\mathbf{x^t}\in\mathbb{R}^{M\times C}$, which is then fed to the scorer network and the Top-K operator (Sec.~\ref{sec:dtsn}) to generate the indicator matrix of the frames with top $K$ highest scores: $\mathbf{Y^{t}}\in\mathbb{R}^{M \times K}$. Finally, we reshape the input $\mathbf{x}$ to $\mathbf{\overline{x}} \in \mathbb{R}^{M \times (N \times C)}$ and extract the selected $K$ frames using the indicator matrix:
\begin{equation}
    \mathbf{\overline{z}} = \mathbf{Y^{t}}^\mathrm{T} \mathbf{\overline{x}} \; \in \mathbb{R}^{K \times (N \times C).}
\end{equation}
The selected tokens are reshaped to $\mathbf{z}\in \mathbb{R}^{K\times N \times C}$ for downstream processing.

\paragraph{\bf Spatial selection.} In contrast to temporal selection, spatial selection is performed on each frame separately, and we aim to select $K$ of the $N$ tokens for each frame.
More specifically, we first feed the tokens of the $m$th frame $\mathbf{x}_m \in \mathbb{R}^{N \times C}$ to the scorer network to generate the importance scores $\mathbf{s}_m\in \mathbb{R}^{N}$. We omit the subscript $m$ for brevity since the same operations are applied to all frames. To obtain the top $K$ spatial tokens, a naive approach is to apply the Top-K operator (Sec.~\ref{sec:dtsn}) directly to the token-based scores $\mathbf{s}$. However, this design inevitably breaks the spatial structure of the input tokens, which is especially inappropriate for spatial selection in video transformers. First, recent top-performing video transformers~\cite{fan2021multiscale,liu2021swin} involve a hierarchical architecture which gradually decreases the spatial resolutions in multiple stages. The discontinuous dropping behavior of tokens is detrimental to local operations like convolutions and pooling for spatial down-sampling. Second, the misalignment of spatial tokens along the temporal dimension makes the temporal modeling~\cite{gberta_2021_ICML} much more challenging, if not impossible. We provide empirical evidence in Table~\ref{tab:spatial} to support our claim.

Instead of performing token-based selection, we introduce a novel anchor-based design for spatial selection. After obtaining the importance scores $\mathbf{s} \in \mathbb{R}^{N}$ for each frame, we first reshape it to a 2D score map $\mathbf{s^s} \in \mathbb{R}^{\sqrt{N} \times \sqrt{N}}$ and split the map into a grid of overlapping anchors $\mathbf{\widetilde{s}^s} \in \mathbb{R}^{G \times K}$ with anchor size $K$, where $G=(\frac{\sqrt{N}-\sqrt{K}}{\alpha} + 1)^2$ is the number of anchors and $\alpha$ is the stride.
Visual examples of the anchors can be found in Figure~\ref{fig:teaser} and~\ref{fig:visualize}.
After that, we aggregate the scores within each anchor via average pooling to obtain the anchor-based scores $\mathbf{s^a} \in \mathbb{R}^{G}$. The original Top-K selection is now cast to a Top-1 selection problem, and we again leverage the Top-K operator (with $K=1$) in Sec.~\ref{sec:dtsn} to obtain the indicator matrix and finally extract the anchor with the highest score.

\section{Experiments}
\label{sec:exp}
In this section, we evaluate the effectiveness of \system by conducting extensive experiments on two large-scale video recognition datasets using two recent video transformer backbones. We introduce the experimental setup in Sec.~\ref{sec:setup}, present main results in Sec.~\ref{sec:results}, and conduct ablation studies to validate the impact of different components in Sec.~\ref{sec:ablation}.

\subsection{Experimental Setup}
\label{sec:setup}
\paragraph{\bf Dataset and backbone.} We mainly use MViT-B16~\cite{fan2021multiscale}, a state-of-the-art video transformer, as the base model and evaluate the effectiveness of \system on Kinetics-400~\cite{kay2017kinetics}. To demonstrate that our approach is generic for different transformer architectures and datasets, we also experiment with the Video Swin Transformer~\cite{liu2021video} on both Kinetics-400 and Something-Something-V2 (SSV2)~\cite{goyal2017something}. Specifically, Kinetics-400 consists of ~240k training videos and 20k validation videos belonging to 400 action categories. SSV2 is a temporally-sensitive dataset, containing 220,847 videos covering 174 action classes. 

We represent different variants of \system by  B-$\text{T}^{L}_{R}$-$\text{S}^{L}_{R}$, where B indicates the backbone network, $\text{T}$ and $\text{S}$ represent the token selection performed along the temporal and spatial dimension, respectively. $L$ and $R$ denote the position where the token selection modules are inserted and the corresponding ratios of selected tokens. For example, MViT-B16-$\text{T}^0_{0.4}$-$\text{S}^4_{0.6}$ indicates performing temporal token selection before the 0th self-attention block with a selection ratio of $0.4$, and spatial token selection before the 4th block with a ratio of $0.6$, using MViT-B16 as the base model.

\paragraph{\bf Evaluation metrics.}
To measure the classification performance, we report the top-1 accuracy on the validation set. We measure the computational cost with FLOPs (floating-point operations), which is a hardware-independent metric.

\paragraph{\bf Implementation details.} In our experiments, we finetune the pre-trained video transformers with our STTS modules. Parameters in the STTS modules are randomly initialized. We set $\sigma$ in Eqn.~\ref{eq:sigma} to 0.1. The learning rate of the backbone layers is set to be 0.01$\times$ of the one in the STTS modules.

To train the MViT models on Kinetics-400, we follow~\cite{fan2021multiscale} to sample a clip of 16 frames with a temporal stride of 4 and a spatial size of $224 \times 224$. The model is trained with AdamW~\cite{loshchilov2018fixing} for 20 epochs with the first 3 epochs for linear warm-up~\cite{goyal2017accurate}. The initial learning rate is set to $1e^{-4}$ for the dynamic token selection module and $1e^{-6}$ for the backbone network, with a mini-batch size of $16$. The cosine learning rate schedule~\cite{loshchilov2016sgdr} is adopted. For the implementation of Video Swin Transformer, we sample a clip of 32 frames from each full-length video using a temporal stride of 2 and spatial size of $224 \times 224$. We set the learning rate of selection networks to $3e^{-4}$, and use 0.01$\times$ learning rate for the backbone model. The batch-size is set to 64. When training on Kinetics-400, we employ an AdamW optimizer~\cite{kingma2014adam} for 10 epochs using a cosine decay learning rate scheduler and 0.8 epochs of linear warm-up. When training on Something-Something-V2, we adopt the AdamW optimizer for longer training of 20 epochs with 0.8 epochs of linear warm-up. For inference, we apply the same testing strategies as the original backbone models for fair comparison. 

\begin{figure}[t]
\begin{minipage}[b]{0.42\textwidth}
  \centering
    \tabcaption{Comparisons of different token selection methods on K400 using the MViT-B16 as backbone.}
  \resizebox{\linewidth}{!}{
  \begin{tabular}{cccccc}
    \toprule
  \textbf{Configs} && \textbf{Rand.} & \textbf{Atten.} & \textbf{GS}  & \textbf{\system} \\
    \cmidrule{1-1} \cmidrule{3-6}
    -$\text{T}^0_{0.5}$ && 75.5 & \light{\textit N/A} & 75.8 & \textbf{77.3} \\ 
    -$\text{T}^4_{0.3}$ && 74.6 & 76.2 & 74.7 & \textbf{77.3} \\ 
    \midrule
    -$\text{S}^0_{0.5}$ && 75.6 & \light{\textit N/A} & \light{\textit N/A} & \textbf{76.2} \\
    -$\text{S}^4_{0.3}$ && 73.6 & 75.6 & \light{\textit N/A} & \textbf{76.8} \\
    \midrule
    -$\text{T}^{0}_{0.8}$-$\text{S}^{0}_{0.7}$ && 75.7 & \light{\textit N/A} & \light{\textit N/A} & \textbf{76.4} \\
    -$\text{T}^{0}_{0.6}$-$\text{S}^{4}_{0.9}$ && 76.4  & \light{\textit N/A} & \light{\textit N/A} & \textbf{77.5} \\
    \bottomrule
  \end{tabular}}
  \label{tab:sel_comparison}
\end{minipage}
 ~
\begin{minipage}[b]{.47\textwidth}
  \centering
  \includegraphics[scale=0.27]{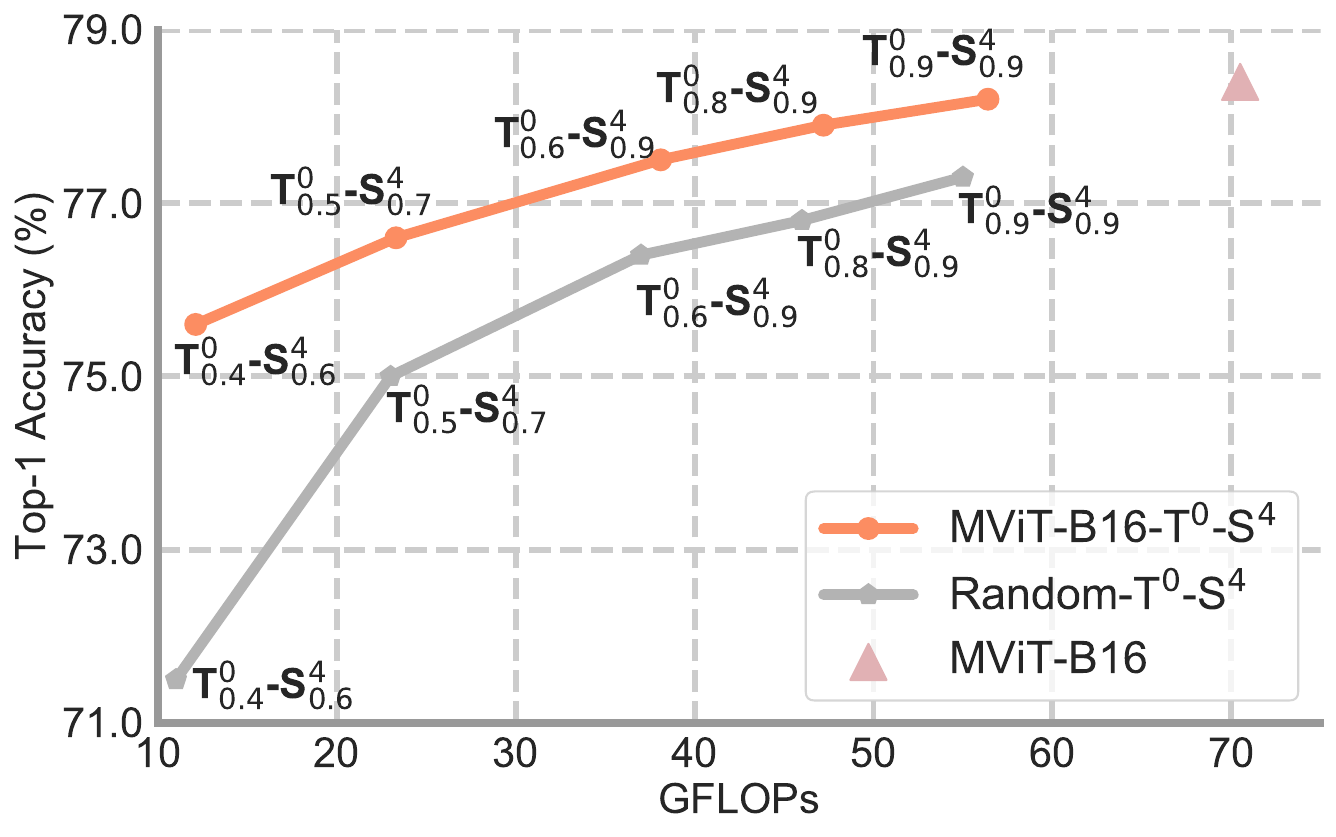}
  \figcaption{Computational cost $vs.$ recognition accuracy on Kinetics-400.} 
  \label{fig:main}
\end{minipage}
\end{figure}

\subsection{Main Results}
\label{sec:results}
\paragraph{\bf Effectiveness of \system.} We first compare \system with some common token selection baselines: 
(1) Random (\textit{Rand.}), which randomly samples $K$ tokens without considering their visual content; 
(2) Attention (\textit{Atten.}), which selects the tokens with the top $K$ highest attention scores with the class token~\footnote{Note that Attention-$K$ cannot be applied if class tokens are not used (\eg, VideoSwin) or self-attention is not yet computed (\eg, 0th block of the model).};
(3) Gumbel-Softmax (\textit{GS}), which uses a Gumbel-softmax trick~\cite{jang2016categorical} for token selection. Note that \textit{GS} cannot be applied to spatial token selection due to the existence of spatial downsampling in recent video transformers. Please refer to the supplementary material for more details.


We summarize the results for temporal-only, spatial-only and joint token selection in Table~\ref{tab:sel_comparison}. The unfeasible baseline settings are filled with \light{\textit N/A}. It can be observed that \system achieves the best accuracy compared to all baseline methods under a similar computational budget. In particular, \system outperforms \textit{GS} by a large margin, although they share the same design of the scorer network (Sec.~\ref{sec:dtsn}), indicating the effectiveness of our differentiable Top-K operator for dynamic token selection.

We further compare our \system with the \textit{Rand.} baseline under different computational budgets. As shown in Figure~\ref{fig:main}, \system consistently achieves superior results, especially for the settings with high reduction rate in computation. For example, MViT-B16-$\text{T}^0_{0.4}$-$\text{S}^4_{0.6}$ outperforms \textit{Rand.} by 5.7\% when using similar 12 GFLOPs. This verifies that informative tokens can be well preserved through our dynamic token selection modules. It can also be observed that the computational overhead of our token selection module is negligible---actually, the parameters and FLOPs of the scorer network are only 1.0\% and 0.7\% of those in the original MViT-B16 backbone.

\begin{table*}[t]
  \centering
  \renewcommand\arraystretch{1.2}
\begin{minipage}[t]{0.47\textwidth}
  \centering
  \tabcaption{Results of \system and comparisons with state-of-the-art methods on  Kinetics-400.}
  \resizebox{\linewidth}{!}{\begin{tabular}{lcccc}
   \toprule
   \multicolumn{1}{c}{\bf Model} && \textbf{Pretrain} & \textbf{TFLOPs}   & \textbf{Top-1} \\
   \cmidrule{1-1} \cmidrule{3-5} 
    X3D-L~\cite{feichtenhofer2020x3d} && - & 0.74  & 77.5 \\
    TimeSformer~\cite{gberta_2021_ICML} && IN-21K & 0.59 & 78.0 \\
    \midrule
    MViT-B16~\cite{fan2021multiscale} && - & 0.35  & 78.4 \\
    \textbf{MViT-B16-$\text{T}^0_{0.8}$-$\text{S}^4_{0.9}$} && - & 0.24 & 77.9 \\
    \textbf{MViT-B16-$\text{T}^0_{0.9}$-$\text{S}^4_{0.9}$} && - & 0.28  & 78.1 \\
    \midrule
    \midrule
    SlowFast 8$\times$8~\cite{feichtenhofer2019slowfast} && - & 3.18 & 77.9 \\
    CorrNet-101~\cite{wang2020video} && - & 6.72 & 79.2 \\
    ViT-B-VTN~\cite{neimark2021video} && IN-21K & 4.22 & 78.6\\
    TimeSformer-HR~\cite{gberta_2021_ICML} && IN-21K & 5.11 & 79.7 \\
    Mformer-HR~\cite{patrick2021keeping} && IN-21K & 28.76 & 81.1 \\
    \midrule 
    VideoSwin-B~\cite{liu2021video} && IN-21K & 3.38 &  82.7 \\
    \textbf{VideoSwin-B-$\text{T}^0_{0.6}$} && IN-21K & 2.17 &  81.4 \\
    \textbf{VideoSwin-B-$\text{T}^0_{0.8}$} && IN-21K & 3.02 &  81.9 \\
    \bottomrule
\end{tabular}}

\label{tab:k400-comparison}		
\end{minipage}
\quad
  \begin{minipage}[t]{.47\textwidth}
  \renewcommand{\arraystretch}{1.25}
    \tabcaption{Results of \system and comparisons with state-of-the-art methods on Something-Something-V2.}
  \resizebox{\linewidth}{!}{\begin{tabular}{lcccc}
    \toprule
 \multicolumn{1}{c}{\bf Model} && \textbf{Pretrain} & \textbf{TFLOPs} & \textbf{Top-1}  \\
    \cmidrule{1-1} \cmidrule{3-5}
    
    TSM~\cite{lin2019tsm} && K400 & 0.37 & 63.3 \\
    STM~\cite{jiang2019stm} && IN-1K & 2.00 & 64.2 \\
    TEA~\cite{li2020tea} && IN-1K & 2.10 &  65.1 \\
    blVNet~\cite{fan2019blvnet} &&  IN-1K & 3.86 &  65.2 \\
    SlowFast 8$\times$8~\cite{feichtenhofer2019slowfast} && K400  & 0.32 &  63.1 \\
    \midrule
    TimeSformer-HR~\cite{gberta_2021_ICML} && IN-21K & 5.11  & 62.5  \\
    ViViT-L~\cite{arnab2021vivit} && K400 & 47.90 & 65.4 \\
    MViT-B16~\cite{fan2021multiscale} && K400 & 0.51 & 67.1 \\
    Mformer-HR~\cite{patrick2021keeping} && K400 & 2.88 &  67.1 \\
    \midrule
    VideoSwin-B~\cite{liu2021video} && K400 & 0.96  & 69.6 \\
    \textbf{VideoSwin-B-$\text{T}^0_{0.6}$} && K400 & 0.57  & 68.1 \\
     \textbf{VideoSwin-B-$\text{T}^0_{0.7}$} && K400 & 0.71 & 68.7 \\
    \bottomrule
  \end{tabular}}
  \label{tab:sthv2-comparison}
  \end{minipage}
\end{table*}

\paragraph{\bf Comparison with the state of the art.} In Table~\ref{tab:k400-comparison}, we compare \system with state-of-the-art video recognition models on Kinetics-400~\cite{carreira2017quo}, including both CNN-based and Transformer-based models. To demonstrate that our approach can be generalized to different transformer architectures, we use both MViT~\cite{fan2021multiscale} and  VideoSwin~\cite{liu2021video} as the base models. We report the overall computational cost at inference---the cost for a single view $\times$ the number of views in space and time, given in Tera-FLOPs (TFLOPs). For clear comparison, we separate the models into two groups and compare \system with those with comparable TFLOPs. Note that the default settings of our STTS are MViT-$\text{T}^{0}$-$\text{S}^{4}$ and VideoSwin-B-$\text{T}^0$ as they achieve the most competitive results according to the ablation study, and we will explore other options in Sec.~\ref{sec:ablation}. Furthermore, we only perform temporal selection for VideoSwin-B, since window shuffling operations are complicated~\cite{DBLP:journals/corr/abs-2105-12723} in the Swin transformer, which makes the spatial selection particularly challenging. It has been recently shown that the shifting operation might not be necessary in modern architectures~\cite{yang2021focal}.

We observe that models equipped with our \system exhibit favorable complexity/accuracy trade-offs at the two complexity levels. Notably, our MViT-B16-$\text{T}^0_{0.9}$-$\text{S}^4_{0.9}$ achieves comparable results with the recent TimeSformer~\cite{gberta_2021_ICML} while requiring only 47\% of the computational cost. Similarly, our VideoSwin-B-$\text{T}^0_{0.6}$ outperforms the recent video transformers (\eg, TimeSformer-HR~\cite{gberta_2021_ICML}, Mformer-HR~\cite{patrick2021keeping}) while saving more than 50\% in computation.
It is also worth mentioning that for both two base models, our \system is capable of saving at least $10\%$ of the computational cost with less than $1\%$ accuracy drop.

We also conduct experiments on Something-Something-V2~\cite{goyal2017something} in Table~\ref{tab:sthv2-comparison}. Given that the pretrained models on MViT~\cite{fan2021multiscale} are not available on SSV2, we only evaluate \system using the VideoSwin Transformer as backbone. We observe that \system outperforms recent transformer models by a large margin with much less computational cost. For example, VideoSwin-B-$\text{T}^0_{0.6}$ achieves an accuracy of $68.1\%$, $2.7\%$ / $1.0\%$ higher than ViViT-L~\cite{arnab2021vivit} / Mformer-HR~\cite{patrick2021keeping}.
We would like to point out that action recognition on SSV2 relies heavily on temporal information, and we believe that the competitive results of our VideoSwin-B-$\text{T}^0_{0.7}$ again verifies the effectiveness of our \system modules in selecting salient temporal tokens.

\begin{table*}[t]
  \centering
  \renewcommand\arraystretch{1.2}
\begin{minipage}[t]{0.36\textwidth}
  \centering
   \caption{Ablation on token selection at different locations.}
  \resizebox{0.85\linewidth}{!}{\begin{tabular}{cccc}
   \toprule
     \textbf{Config} && \textbf{GFLOPs}  & \textbf{Top-1} \\
     \cmidrule{1-1} \cmidrule{3-4}
    MViT-B16 && 70.5 & 78.4 \\
    \midrule
    \midrule
     -$\text{S}^{0}_{0.6}$ && 31.2 (\Drop{55.7\%}) & 76.2 \\
     -$\text{S}^{4}_{0.3}$ && 35.2 (\Drop{50.1\%}) & 76.8 \\
     \midrule
      -$\text{T}^{0}_{0.6}$  && 41.3 (\Drop{41.4\%}) & 77.5 \\
      -$\text{T}^{4}_{0.3}$ && 37.5 (\Drop{46.8\%}) & 77.3 \\
     \midrule
      -$\text{T}^{0}_{0.8}$-$\text{S}^{0}_{0.7}$ && 36.0 (\Drop{48.9\%}) & 76.4 \\
      -$\text{T}^{0}_{0.6}$-$\text{S}^{4}_{0.9}$ && 38.1 (\Drop{46.0\%}) & \textbf{77.5} \\
     \bottomrule
  \end{tabular}}
   \label{tab:st_joint}	
\end{minipage}
~
  \begin{minipage}[t]{.6\textwidth}
  \renewcommand\arraystretch{1.2}
    \caption{Multi-step token selection on Kinetics-400 using MViT as the base model. The savings in GFLOPs are all around 50\%.}
  \resizebox{\linewidth}{!}{
  \begin{tabular}{ccccccccc}
    \toprule
    \multicolumn{2}{c}{\textbf{Temporal}} && \multicolumn{2}{c}{\textbf{Spatial}} &&
    \multirow{2}{*}{\textbf{GFLOPs}} & \multirow{2}{*}{\textbf{Top-1}} \\
    Location & Ratio && Location & Ratio && ~ & ~\\
    \cmidrule{1-2} \cmidrule{4-5} \cmidrule{7-8}
    0, 4 & 0.9, 0.7 && 0, 4 & 0.9, 0.7 && 35.8 (\Drop{49.2\%}) & 76.4 \\
    4, 8 & 0.6, 0.4 && 4, 8 & 0.7, 0.7 && 36.0 (\Drop{48.9\%} & 76.5 \\
    0, 4, 8 & 0.9, 0.9, 0.7 && 0, 4, 8 & 0.9, 0.9, 0.8 && 36.6 (\Drop{48.1\%}) & 76.6 \\
    \midrule
    0 & 0.6 && 4 & 0.9 &&  37.5 (\Drop{46.8\%}) & \textbf{77.5} \\
    \bottomrule
  \end{tabular}}

  \label{tab:progressive}
  \end{minipage}
\end{table*}

\subsection{Discussion}
\label{sec:ablation}
\paragraph{\bf Token selection at different locations / multiple steps.} The flexibility of our \system module implies that different choices of token selection configuration are available in order to achieve a similar computational reduction rate. For example, in order to reduce the computation of MViT-B16 by approximately 50\%, one can either apply (1) a spatial-only / temporal-only token selection at early stages with a higher selection ratio (\eg, -$\text{S}^{0}_{0.6}$ / -$\text{T}^{0}_{0.6}$); (2) a spatial-only / temporal-only token selection at later stages with a lower selection ratio  (\eg, -$\text{S}^{4}_{0.3}$ / -$\text{T}^{4}_{0.3}$); or (3) a joint token selection (\eg, -$\text{T}^{0}_{0.6}$-$\text{S}^{4}_{0.9}$), optionally in a multi-step manner (Table~\ref{tab:progressive}). We provide an in-depth analysis of these choices in this section, taking MViT-B16 as an example and evaluating on Kinetics-400. We report the inference cost for a single view.

Table~\ref{tab:st_joint} shows the performance of \system when token selection is performed at different locations. For all the settings, \emph{the ratios of selected tokens are adjusted to ensure the overall computational cost is reduced by approximately 50\%}. We observe that temporal selection exceeds the spatial selection by a large margin, which demonstrates that temporal redundancy is more significant than spatial redundancy in videos. In addition, joint token selection (with temporal token selection at the early layer and spatial token selection at the deep layer, \ie, -$\text{T}^{0}_{0.6}$-$\text{S}^{4}_{0.9}$) achieves the best result.
We also perform the token selection in a multi-step manner, \ie, performing multiple token selections at different layers of a transformer network with a higher selection ratio for each of them. The results in Table~\ref{tab:progressive} shows that although sharing a similar computational cost, multi-step selection produces inferior results than the single-step selection. We hypothesize that the multi-step selection leads to frequent changes of the spatio-temporal structures of the videos and is more difficult to train.


\paragraph{\bf Impact of anchor-based spatial selection.} To verify the effectiveness of our anchor-based spatial selection described in Sec.~\ref{sec:tsts}, we compare different spatial selection strategies in Table~\ref{tab:spatial}. Specifically, \Checkmark in the second column indicates using the anchor-based spatial selection while \XSolidBrush indicates using the token-based selection. T and A in the third column denote taking token-level features or anchor-level features (average-pooled features within each anchor) as input for computing the importance scores, respectively. 

We observe that the naive token-based spatial selection results in a remarkable performance drop (76.2\% $\rightarrow$ 57.6\%). It is clear that such a spatial selection strategy breaks the spatial structure of the original backbone model and therefore the \textit{Rand.} baseline performs particularly poorly (16.7\%) without finetuning model parameters. We also observe that taking anchor-level features as input for scorer network performs worse than our default setting.

\begin{table*}[t]
  \centering
  \begin{minipage}[t]{0.36\textwidth}
  \renewcommand\arraystretch{1.1}
    \caption{Impact of anchor-based spatial selection.}
  \resizebox{\linewidth}{!}{
  \begin{tabular}{ccccc}
    \toprule
    \textbf{Config} && \textbf{Anchor} & \textbf{Score} & \textbf{Top-1} \\
    \cmidrule{1-1} \cmidrule{3-5}
    \multirow{2}{*}{\textit{Rand.}-$\text{S}^{0}_{0.5}$} && \XSolidBrush & - & 16.7 \\
      && \Checkmark & - & 75.6 \\
     \midrule
       && \XSolidBrush & T & 57.6 \\
     MViT-B16-$\text{S}^{0}_{0.5}$  && \Checkmark  & A & 75.4 \\
      ~ && \Checkmark  & T & \textbf{76.2} \\
    \bottomrule
 \end{tabular}}
\label{tab:spatial}	
\end{minipage}
~
  \begin{minipage}[t]{.33\textwidth}
  \renewcommand\arraystretch{1.2}
    \caption{Comparison of the inference time.}
  \resizebox{\linewidth}{!}{
  \begin{tabular}{cccc}
    \toprule
    \multirow{2}{*}{\textbf{Config}} && \multirow{2}{*}{\textbf{GFLOPs}} & \textbf{Throughput} \\
    ~ && ~ & (video / s)  \\
    \cmidrule{1-1} \cmidrule{3-4}
     MViT-B16 && 70.5 & 79.3 \\
     \midrule
     -$\text{T}^0_{0.9}$-$\text{S}^4_{0.9}$ && 56.4 & 96.5 \\
     -$\text{T}^0_{0.8}$-$\text{S}^4_{0.9}$ && 47.2 & 112.9 \\
     -$\text{T}^0_{0.6}$-$\text{S}^4_{0.9}$ && 38.1 & 129.7 \\
    \bottomrule
  \end{tabular}}
\label{tab:inference}
  \end{minipage}
~
  \begin{minipage}[t]{0.25\textwidth}
  \renewcommand\arraystretch{1.15}
   \centering
    \caption{Impact of $\sigma$ values.} 
   \resizebox{0.9\linewidth}{!}{
   \begin{tabular}{cccc}
    \toprule
    \textbf{Config} && \textbf{$\sigma$} & \textbf{Top-1} \\
    \cmidrule{1-1} \cmidrule{3-4}
     Rand. && - & 75.5 \\
     \midrule
     -$\text{T}^0_{0.5}$ && 0.05 & 77.2 \\
     -$\text{T}^0_{0.5}$ && 0.1 & \textbf{77.3} \\
     -$\text{T}^0_{0.5}$ && 0.2 & 77.1 \\
    \bottomrule
  \end{tabular}}
\label{tab:sigma}		
 \end{minipage}
\end{table*}

\paragraph{\bf Inference time.} To verify that our method effectively reduces the computational overhead, we measure the inference time of \system and MViT-B16 on a single 8 RTX 3090 GPU server. The comparison results are reported in Table~\ref{tab:inference}, which illustrate that \system indeed reduces the inference time in practice.

\begin{figure*}[t]
  \centering
  \includegraphics[width=\linewidth]{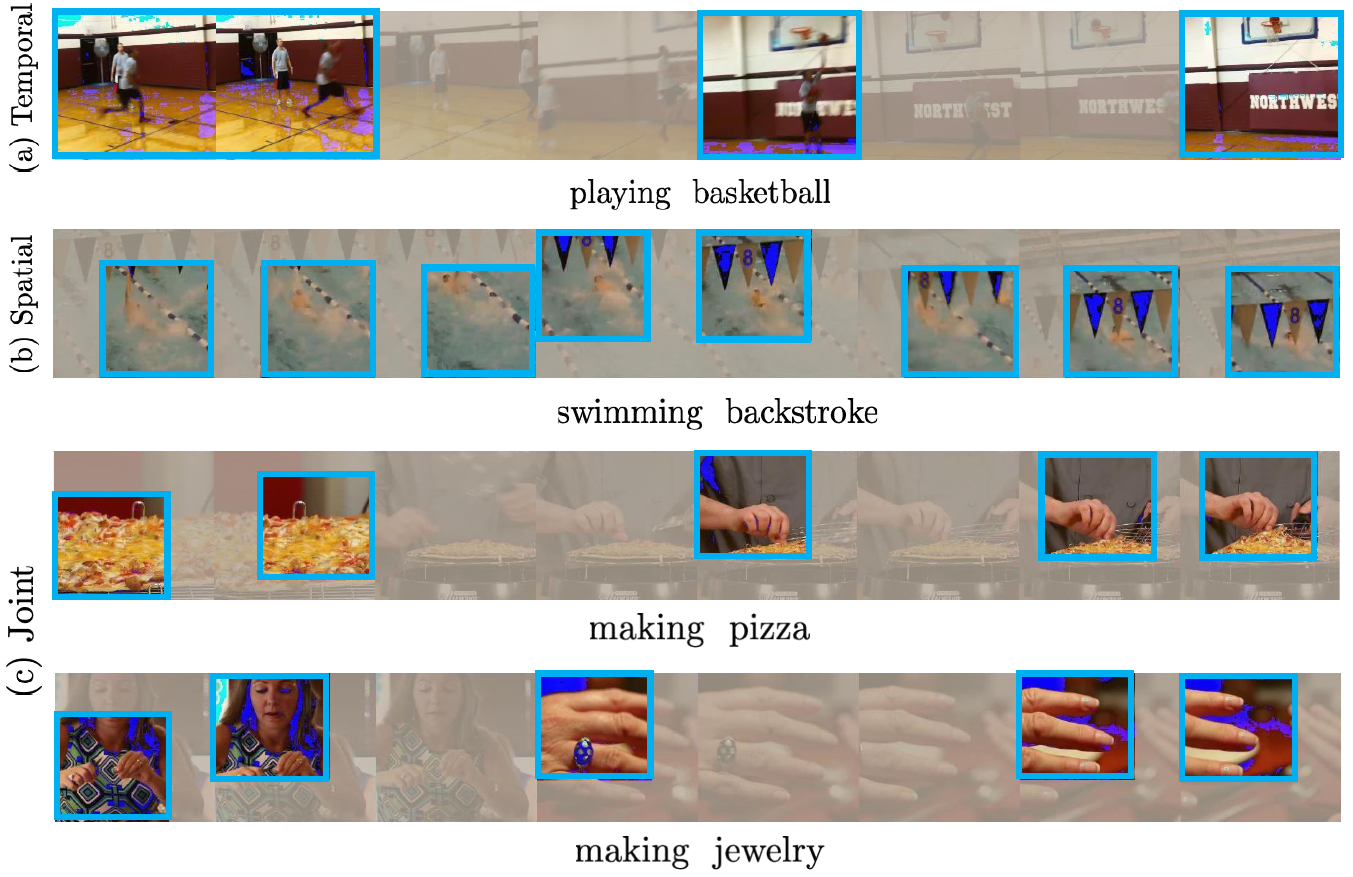}
  \caption{Visualization of the temporal, spatial, and joint token selection results by our \system, where the masked frames / regions are pruned and only the tokens inside the bounding box will be fed to the subsequent video transformers. }
  \label{fig:visualize}
\end{figure*}


\paragraph{\bf Impact of $\sigma$.} To investigate the influence of $\sigma$ in Eqn.~\ref{eq:sigma}, we train MViT-B16-$\text{T}^0_{0.5}$ using different $\sigma$ values and report the comparison results in Table~\ref{tab:sigma}. We can see that \system is insensitive to the choice of hyper-parameters, and it consistently outperforms the \textit{Rand.} baseline. Unless otherwise mentioned, we set $\sigma=0.1$ in our experiment since it achieves slightly better performance than other settings.

\paragraph{\bf Qualitative results.}
We visualize the temporal-only, spatial-only, and joint token selection results in Figure~\ref{fig:visualize}, where the masked frames and regions are discarded by \system. We observe that our \system can not only effectively identify the most informative frames in a video clip, but also locate the discriminative regions inside each frame. With such token pruning in temporal and spatial dimensions, only the tokens that are important for correctly recognizing the actions are retained and fed into the subsequent video transformers, resulting in a saved computational cost with minimal drop of classification accuracy.

\section{Conclusion}
\label{sec:conclusion}
In this paper, we introduced \system, a dynamic spatio-temporal token selection framework, to reduce both temporal and spatial redundancies of video transformers for efficient video recognition. We formulated the token selection as a ranking problem, where the importance of different tokens was predicted by a lightweight selection network and only those with Top-K scores will be preserved for subsequent processing. In the temporal dimension, we selected a subset of frames with the highest relevance to the action category, while in the spatial dimension, we retained the most discriminative region within each frame to preserve the structural information of the input frame. To enable the end-to-end training of the backbone model together with the token selection module, we leveraged a perturbed-maximum based differentiable Top-K operator. Extensive experiments on several video recognition benchmarks validated our \system could achieve competitive efficiency-accuracy trade-offs. 
\newline
\\
\textbf{Acknowledgement} Y.-G. Jiang was
sponsored in part by ``Shuguang Program'' supported by Shanghai Education Development Foundation and Shanghai Municipal
Education Commission (No. 20SG01). Z. Wu was supported by NSFC under Grant No. 62102092.

\clearpage
%
%
\bibliographystyle{splncs04}
\bibliography{main}

\begin{thebibliography}{10}
\providecommand{\url}[1]{\texttt{#1}}
\providecommand{\urlprefix}{URL }
\providecommand{\doi}[1]{https://doi.org/#1}

\bibitem{abernethy2016perturbation}
Abernethy, J., Lee, C., Tewari, A.: Perturbation techniques in online learning
  and optimization. Perturbations, Optimization, and Statistics  (2016)

\bibitem{arnab2021vivit}
Arnab, A., Dehghani, M., Heigold, G., Sun, C., Lu{\v{c}}i{\'c}, M., Schmid, C.:
  Vivit: A video vision transformer. In: ICCV (2021)

\bibitem{gberta_2021_ICML}
Bertasius, G., Wang, H., Torresani, L.: Is space-time attention all you need
  for video understanding? In: ICML (2021)

\bibitem{berthet2020learning}
Berthet, Q., Blondel, M., Teboul, O., Cuturi, M., Vert, J.P., Bach, F.:
  Learning with differentiable perturbed optimizers. arXiv preprint
  arXiv:2002.08676  (2020)

\bibitem{bhardwaj2019efficient}
Bhardwaj, S., Srinivasan, M., Khapra, M.M.: Efficient video classification
  using fewer frames. In: CVPR (2019)

\bibitem{carion2020end}
Carion, N., Massa, F., Synnaeve, G., Usunier, N., Kirillov, A., Zagoruyko, S.:
  End-to-end object detection with transformers. In: ECCV (2020)

\bibitem{carreira2017quo}
Carreira, J., Zisserman, A.: Quo vadis, action recognition? a new model and the
  kinetics dataset. In: CVPR (2017)

\bibitem{cordonnier2021differentiable}
Cordonnier, J.B., Mahendran, A., Dosovitskiy, A., Weissenborn, D., Uszkoreit,
  J., Unterthiner, T.: Differentiable patch selection for image recognition.
  In: CVPR (2021)

\bibitem{cuturi2019differentiable}
Cuturi, M., Teboul, O., Vert, J.P.: Differentiable ranking and sorting using
  optimal transport. In: NeurIPS (2019)

\bibitem{davidson2010youtube}
Davidson, J., Liebald, B., Liu, J., Nandy, P., Van~Vleet, T., Gargi, U., Gupta,
  S., He, Y., Lambert, M., Livingston, B., et~al.: The youtube video
  recommendation system. In: RS (2010)

\bibitem{devlin2018bert}
Devlin, J., Chang, M.W., Lee, K., Toutanova, K.: Bert: Pre-training of deep
  bidirectional transformers for language understanding. arXiv preprint
  arXiv:1810.04805  (2018)

\bibitem{dong2019dual}
Dong, J., Li, X., Xu, C., Ji, S., He, Y., Yang, G., Wang, X.: Dual encoding for
  zero-example video retrieval. In: CVPR (2019)

\bibitem{dosovitskiy2021an}
Dosovitskiy, A., Beyer, L., Kolesnikov, A., Weissenborn, D., Zhai, X.,
  Unterthiner, T., Dehghani, M., Minderer, M., Heigold, G., Gelly, S.,
  Uszkoreit, J., Houlsby, N.: An image is worth 16x16 words: Transformers for
  image recognition at scale. In: ICLR (2021)

\bibitem{fan2021multiscale}
Fan, H., Xiong, B., Mangalam, K., Li, Y., Yan, Z., Malik, J., Feichtenhofer,
  C.: Multiscale vision transformers. In: ICCV (2021)

\bibitem{fan2019blvnet}
Fan, Q., Chen, C.F.R., Kuehne, H., Pistoia, M., Cox, D.: {More Is Less:
  Learning Efficient Video Representations by Temporal Aggregation Modules}.
  In: NeurIPS (2019)

\bibitem{feichtenhofer2020x3d}
Feichtenhofer, C.: X3d: Expanding architectures for efficient video
  recognition. In: CVPR (2020)

\bibitem{feichtenhofer2019slowfast}
Feichtenhofer, C., Fan, H., Malik, J., He, K.: Slowfast networks for video
  recognition. In: ICCV (2019)

\bibitem{feichtenhofer2016convolutional}
Feichtenhofer, C., Pinz, A., Zisserman, A.: Convolutional two-stream network
  fusion for video action recognition. In: CVPR (2016)

\bibitem{gabeur2020multi}
Gabeur, V., Sun, C., Alahari, K., Schmid, C.: Multi-modal transformer for video
  retrieval. In: ECCV (2020)

\bibitem{goyal2017accurate}
Goyal, P., Doll{\'a}r, P., Girshick, R., Noordhuis, P., Wesolowski, L., Kyrola,
  A., Tulloch, A., Jia, Y., He, K.: Accurate, large minibatch sgd: Training
  imagenet in 1 hour. arXiv preprint arXiv:1706.02677  (2017)

\bibitem{goyal2017something}
Goyal, R., Ebrahimi~Kahou, S., Michalski, V., Materzynska, J., Westphal, S.,
  Kim, H., Haenel, V., Fruend, I., Yianilos, P., Mueller-Freitag, M., et~al.:
  The" something something" video database for learning and evaluating visual
  common sense. In: ICCV (2017)

\bibitem{hara2018can}
Hara, K., Kataoka, H., Satoh, Y.: Can spatiotemporal 3d cnns retrace the
  history of 2d cnns and imagenet? In: CVPR (2018)

\bibitem{he2020gta}
He, B., Yang, X., Wu, Z., Chen, H., Lim, S.N., Shrivastava, A.: Gta: Global
  temporal attention for video action understanding. In: BMVC (2021)

\bibitem{heo2021rethinking}
Heo, B., Yun, S., Han, D., Chun, S., Choe, J., Oh, S.J.: Rethinking spatial
  dimensions of vision transformers. arXiv preprint arXiv:2103.16302  (2021)

\bibitem{huang2016real}
Huang, Y., Cui, B., Jiang, J., Hong, K., Zhang, W., Xie, Y.: Real-time video
  recommendation exploration. In: ICMD (2016)

\bibitem{jang2016categorical}
Jang, E., Gu, S., Poole, B.: Categorical reparameterization with
  gumbel-softmax. arXiv preprint arXiv:1611.01144  (2016)

\bibitem{jiang2019stm}
Jiang, B., Wang, M., Gan, W., Wu, W., Yan, J.: Stm: Spatiotemporal and motion
  encoding for action recognition. In: ICCV (2019)

\bibitem{kay2017kinetics}
Kay, W., Carreira, J., Simonyan, K., Zhang, B., Hillier, C., Vijayanarasimhan,
  S., Viola, F., Green, T., Back, T., Natsev, P., et~al.: The kinetics human
  action video dataset. arXiv preprint arXiv:1705.06950  (2017)

\bibitem{kingma2014adam}
Kingma, D.P., Ba, J.: Adam: A method for stochastic optimization. arXiv
  preprint arXiv:1412.6980  (2014)

\bibitem{kitaev2020reformer}
Kitaev, N., Kaiser, L., Levskaya, A.: Reformer: The efficient transformer. In:
  ICLR (2020)

\bibitem{kondratyuk2021movinets}
Kondratyuk, D., Yuan, L., Li, Y., Zhang, L., Tan, M., Brown, M., Gong, B.:
  Movinets: Mobile video networks for efficient video recognition. In: CVPR
  (2021)

\bibitem{korbar2019scsampler}
Korbar, B., Tran, D., Torresani, L.: Scsampler: Sampling salient clips from
  video for efficient action recognition. In: ICCV (2019)

\bibitem{lee2017large}
Lee, J., Abu-El-Haija, S.: Large-scale content-only video recommendation. In:
  ICCVW (2017)

\bibitem{li2022uniformer}
Li, K., Wang, Y., Peng, G., Song, G., Liu, Y., Li, H., Qiao, Y.: Uniformer:
  Unified transformer for efficient spatial-temporal representation learning.
  In: ICLR (2022)

\bibitem{li2021uav}
Li, T., Liu, J., Zhang, W., Ni, Y., Wang, W., Li, Z.: Uav-human: A large
  benchmark for human behavior understanding with unmanned aerial vehicles. In:
  CVPR (2021)

\bibitem{li2020tea}
Li, Y., Ji, B., Shi, X., Zhang, J., Kang, B., Wang, L.: Tea: Temporal
  excitation and aggregation for action recognition. In: CVPR (2020)

\bibitem{lin2019tsm}
Lin, J., Gan, C., Han, S.: Tsm: Temporal shift module for efficient video
  understanding. In: ICCV (2019)

\bibitem{liu2021swin}
Liu, Z., Lin, Y., Cao, Y., Hu, H., Wei, Y., Zhang, Z., Lin, S., Guo, B.: Swin
  transformer: Hierarchical vision transformer using shifted windows. In: ICCV
  (2021)

\bibitem{liu2021video}
Liu, Z., Ning, J., Cao, Y., Wei, Y., Zhang, Z., Lin, S., Hu, H.: Video swin
  transformer. arXiv preprint arXiv:2106.13230  (2021)

\bibitem{liu2020teinet}
Liu, Z., Luo, D., Wang, Y., Wang, L., Tai, Y., Wang, C., Li, J., Huang, F., Lu,
  T.: Teinet: Towards an efficient architecture for video recognition. In: AAAI
  (2020)

\bibitem{loshchilov2016sgdr}
Loshchilov, I., Hutter, F.: Sgdr: Stochastic gradient descent with warm
  restarts. arXiv preprint arXiv:1608.03983  (2016)

\bibitem{loshchilov2018fixing}
Loshchilov, I., Hutter, F.: Fixing weight decay regularization in adam  (2018)

\bibitem{mei2011contextual}
Mei, T., Yang, B., Hua, X.S., Li, S.: Contextual video recommendation by
  multimodal relevance and user feedback. TOIS  (2011)

\bibitem{naseer2021intriguing}
Naseer, M., Ranasinghe, K., Khan, S., Hayat, M., Khan, F., Yang, M.H.:
  Intriguing properties of vision transformers. In: NeurIPS (2021)

\bibitem{neimark2021video}
Neimark, D., Bar, O., Zohar, M., Asselmann, D.: Video transformer network.
  arXiv preprint arXiv:2102.00719  (2021)

\bibitem{pan2021iared}
Pan, B., Panda, R., Jiang, Y., Wang, Z., Feris, R., Oliva, A.:
  {IA}-{RED}\${\textasciicircum}2\$: Interpretability-aware redundancy
  reduction for vision transformers. In: NeurIPS (2021)

\bibitem{paszke2019pytorch}
Paszke, A., Gross, S., Massa, F., Lerer, A., Bradbury, J., Chanan, G., Killeen,
  T., Lin, Z., Gimelshein, N., Antiga, L., et~al.: Pytorch: An imperative
  style, high-performance deep learning library. In: NeurIPS (2019)

\bibitem{patrick2021keeping}
Patrick, M., Campbell, D., Asano, Y., Misra, I., Metze, F., Feichtenhofer, C.,
  Vedaldi, A., Henriques, J.F.: Keeping your eye on the ball: Trajectory
  attention in video transformers. In: NeurIPS (2021)

\bibitem{rao2021dynamicvit}
Rao, Y., Zhao, W., Liu, B., Lu, J., Zhou, J., Hsieh, C.J.: Dynamicvit:
  Efficient vision transformers with dynamic token sparsification. In: NeurIPS
  (2021)

\bibitem{ryoo2021tokenlearner}
Ryoo, M.S., Piergiovanni, A., Arnab, A., Dehghani, M., Angelova, A.:
  Tokenlearner: Adaptive space-time tokenization for videos. In: NeurIPS (2021)

\bibitem{sun2022human}
Sun, Z., Ke, Q., Rahmani, H., Bennamoun, M., Wang, G., Liu, J.: Human action
  recognition from various data modalities: A review. IEEE TPAMI  (2022)

\bibitem{touvron2021training}
Touvron, H., Cord, M., Douze, M., Massa, F., Sablayrolles, A., J{\'e}gou, H.:
  Training data-efficient image transformers \& distillation through attention.
  In: ICML (2021)

\bibitem{tran2018closer}
Tran, D., Wang, H., Torresani, L., Ray, J., LeCun, Y., Paluri, M.: A closer
  look at spatiotemporal convolutions for action recognition. In: CVPR (2018)

\bibitem{vaswani2017attention}
Vaswani, A., Shazeer, N., Parmar, N., Uszkoreit, J., Jones, L., Gomez, A.N.,
  Kaiser, {\L}., Polosukhin, I.: Attention is all you need. In: NeurIPS (2017)

\bibitem{wang2020video}
Wang, H., Tran, D., Torresani, L., Feiszli, M.: Video modeling with correlation
  networks. In: CVPR (2020)

\bibitem{wang2021tdn}
Wang, L., Tong, Z., Ji, B., Wu, G.: Tdn: Temporal difference networks for
  efficient action recognition. In: CVPR (2021)

\bibitem{wang2022bevt}
Wang, R., Chen, D., Wu, Z., Chen, Y., Dai, X., Liu, M., Jiang, Y.G., Zhou, L.,
  Yuan, L.: Bevt: Bert pretraining of video transformers. In: CVPR (2022)

\bibitem{wang2020linformer}
Wang, S., Li, B.Z., Khabsa, M., Fang, H., Ma, H.: Linformer: Self-attention
  with linear complexity. arXiv preprint arXiv:2006.04768  (2020)

\bibitem{wang2021adaptive}
Wang, Y., Chen, Z., Jiang, H., Song, S., Han, Y., Huang, G.: Adaptive focus for
  efficient video recognition. In: ICCV (2021)

\bibitem{wang2021adafocus}
Wang, Y., Yue, Y., Lin, Y., Jiang, H., Lai, Z., Kulikov, V., Orlov, N., Shi,
  H., Huang, G.: Adafocus v2: End-to-end training of spatial dynamic networks
  for video recognition. In: CVPR (2022)

\bibitem{wang2021end}
Wang, Y., Xu, Z., Wang, X., Shen, C., Cheng, B., Shen, H., Xia, H.: End-to-end
  video instance segmentation with transformers. In: CVPR (2021)

\bibitem{wu2018compressed}
Wu, C.Y., Zaheer, M., Hu, H., Manmatha, R., Smola, A.J., Kr{\"a}henb{\"u}hl,
  P.: Compressed video action recognition. In: CVPR (2018)

\bibitem{wu2020dynamic}
Wu, Z., Li, H., Xiong, C., Jiang, Y.G., Davis, L.S.: A dynamic frame selection
  framework for fast video recognition. IEEE TPAMI  (2022)

\bibitem{DBLP:journals/ijcv/WuLZXJD21}
Wu, Z., Li, H., Zheng, Y., Xiong, C., Jiang, Y., Davis, L.S.: A coarse-to-fine
  framework for resource efficient video recognition. IJCV  (2021)

\bibitem{wu2019adaframe}
Wu, Z., Xiong, C., Ma, C.Y., Socher, R., Davis, L.S.: Adaframe: Adaptive frame
  selection for fast video recognition. In: CVPR (2019)

\bibitem{xie2020differentiable}
Xie, Y., Dai, H., Chen, M., Dai, B., Zhao, T., Zha, H., Wei, W., Pfister, T.:
  Differentiable top-k with optimal transport. In: NeurIPS (2020)

\bibitem{xu2021sutd}
Xu, L., Huang, H., Liu, J.: Sutd-trafficqa: A question answering benchmark and
  an efficient network for video reasoning over traffic events. In: CVPR (2021)

\bibitem{yang2021focal}
Yang, J., Li, C., Zhang, P., Dai, X., Xiao, B., Yuan, L., Gao, J.: Focal
  self-attention for local-global interactions in vision transformers. In:
  NeurIPS (2021)

\bibitem{yeung2016end}
Yeung, S., Russakovsky, O., Mori, G., Fei-Fei, L.: End-to-end learning of
  action detection from frame glimpses in videos. In: CVPR (2016)

\bibitem{yuan2020central}
Yuan, L., Wang, T., Zhang, X., Tay, F.E., Jie, Z., Liu, W., Feng, J.: Central
  similarity quantization for efficient image and video retrieval. In: CVPR
  (2020)

\bibitem{zhang2020feature}
Zhang, D., Zhang, H., Tang, J., Wang, M., Hua, X., Sun, Q.: Feature pyramid
  transformer. In: ECCV (2020)

\bibitem{DBLP:journals/corr/abs-2105-12723}
Zhang, Z., Zhang, H., Zhao, L., Chen, T., Pfister, T.: Aggregating nested
  transformers. In: AAAI (2022)

\bibitem{zheng2021rethinking}
Zheng, S., Lu, J., Zhao, H., Zhu, X., Luo, Z., Wang, Y., Fu, Y., Feng, J.,
  Xiang, T., Torr, P.H., et~al.: Rethinking semantic segmentation from a
  sequence-to-sequence perspective with transformers. In: CVPR (2021)

\bibitem{zheng2020dynamic}
Zheng, Y.D., Liu, Z., Lu, T., Wang, L.: Dynamic sampling networks for efficient
  action recognition in videos. TIP  (2020)

\bibitem{zhu2021deformable}
Zhu, X., Su, W., Lu, L., Li, B., Wang, X., Dai, J.: Deformable {\{}detr{\}}:
  Deformable transformers for end-to-end object detection. In: ICLR (2021)

\bibitem{zolfaghari2018eco}
Zolfaghari, M., Singh, K., Brox, T.: Eco: Efficient convolutional network for
  online video understanding. In: ECCV (2018)

\end{thebibliography}
\end{document}